\documentclass[a4paper]{article}

\usepackage[english]{babel}
\usepackage{amsthm}  
\usepackage{amsmath} 
\usepackage{amssymb} 
\usepackage{mathabx} 
\usepackage{mathtools}
\usepackage{graphicx}
\usepackage{fancyhdr}
\usepackage{setspace}
\usepackage{lineno}	 
\usepackage{lastpage}

\usepackage{subcaption}
\usepackage{adjustbox}
\usepackage{makecell}
\usepackage[numbers,sort&compress]{natbib}	
\usepackage{todonotes}

\usepackage{booktabs}
\usepackage{multirow}

\usepackage{comment} 

\newcommand{\mytitle}{Physical Pooling Functions in Graph Neural Networks for Molecular Property Prediction}
\newcommand{\myshorttitle}{Physical Pooling Functions in Graph Neural Networks}
\newcommand{\myauthor}{Artur M. Schweidtmann$^{1,2}$, Jan G. Rittig$^{1}$, Jana M. Weber$^{3}$, Martin Grohe$^{4}$, Manuel Dahmen$^{5}$, Kai Leonhard$^{6}$, Alexander Mitsos$^{*,7,5,1}$} 
\newcommand{\myauthorshort}{Schweidtmann et al.}

\usepackage[colorlinks,linkcolor=blue,citecolor=blue,urlcolor=blue,pdftitle={\mytitle},pdfauthor={Artur M. Schweidtmann}]{hyperref}

{
\fancyhead[L]{\myshorttitle}
\fancyhead[R]{\small{\textit{\the\day.\the\month.\the\year}}}
\fancyfoot[L]{\copyright \small{\textit{\myauthorshort}}}
\fancyfoot[R]{Page \thepage\ of \pageref*{LastPage}}
}

\fancypagestyle{firststyle}
{
 \fancyhead[L]{\copyright \small{\textit{\myauthorshort}}}
\fancyhead[R]{Page \thepage\ of \pageref*{LastPage}}
\fancyfoot[C]{}
 \fancyfoot[R]{}
}

\addto\captionsenglish{%
}

\DeclareMathOperator{\mean}{mean}
\DeclareMathOperator{\sumWord}{sum}

\newtheoremstyle{myplain}
{} 
{} 
{\itshape} 
{} 
{} 
{} 
{ } 
{\textbf{\thmname{#1}\thmnumber{ #2}}\thmnote{ (#3)}} 

\newtheoremstyle{mydefinition}
{} 
{} 
{} 
{} 
{} 
{} 
{ } 
{\textbf{\thmname{#1}\thmnumber{ #2}}\thmnote{ (\textit{#3})}} 

\newtheoremstyle{myremark}
{} 
{} 
{} 
{} 
{} 
{} 
{ } 
{\textit{\thmname{#1}\thmnumber{ #2}\thmnote{ (#3)}}} 

\newtheoremstyle{mynote}
{} 
{} 
{} 
{} 
{} 
{:} 
{ } 
{\textit{\thmname{#1}\thmnumber{ #2}\thmnote{ (#3)}}} 

\definecolor{rwth}{rgb}{0,0.32,0.62}
\definecolor{rwth-75}{rgb}{0.25,0.49,0.71}
\definecolor{rwth-50}{rgb}{0.55,0.73,0.89}
\definecolor{grun}{rgb}{0.34,0.67,0.15}
\definecolor{rot}{rgb}{0.8,0.02,0.11}
\definecolor{magenta}{RGB}{227,0,102}
\definecolor{petrol}{RGB}{0,97,101}
\definecolor{violett}{RGB}{97,33,88}
\definecolor{maigrun}{RGB}{189,205,0}

\makeatletter
\let\@addpunct\@gobble
\g@addto@macro{\thm@space@setup}{\thm@headpunct{}} 
\makeatother

\renewenvironment{abstract}{\noindent\textbf{Abstract:}}{}

\begin{document}
	
	\thispagestyle{firststyle}
	\begin{flushleft}\begin{large}\textbf{\mytitle}\end{large} \end{flushleft}
	\myauthor

	\begin{flushleft}\begin{small}
			$^1$ RWTH Aachen University, Process Systems Engineering (AVT.SVT), Forckenbeckstr. 51, 52074 Aachen, Germany \\[0.25cm]
			
			$^2$ Current address: Delft University of Technology, Department of Chemical Engineering, Van der Maasweg 9, Delft 2629 HZ, The Netherlands \\[0.25cm]
				
			$^3$ Delft University of Technology, Delft Bioinformatics Lab, Intelligent Systems, TU Delft, 2628 XE Delft, The Netherlands \\[0.25cm]
			
			$^4$ RWTH Aachen University, Lehrstuhl für Informatik 7, Ahornstr. 55, 52074 Aachen, Germany \\[0.25cm]
			
			$^5$ Forschungszentrum Jülich GmbH, Institute of Energy and Climate Research IEK-10 -- Energy Systems Engineering, Wilhelm-Johnen-Str., 52425 Jülich, Germany \\[0.25cm]
			
			$^6$ RWTH Aachen University, Institute of Technical Thermodynamics, Schinkelstr. 8, 52062 Aachen, Germany \\[0.25cm]
			
			$^7$ JARA Center for Simulation and Data Science (CSD), Aachen, Germany \\[0.25cm]
			
			$^*$ Corresponding author: {amitsos@alum.mit.edu}

			%

			%
			%
			%
			%
			%

		\end{small}
	\end{flushleft}

%
%
%

%
%

%
%


\begin{abstract}
    Graph neural networks (GNNs) are emerging in chemical engineering for the end-to-end learning of physicochemical properties based on molecular graphs. 
    A key element of GNNs is the pooling function which combines atom feature vectors into molecular fingerprints.
    Most previous works use a standard pooling function to predict a variety of properties. 
    However, unsuitable pooling functions can lead to unphysical GNNs that poorly generalize. 
    We compare and select meaningful GNN pooling methods based on physical knowledge about the learned properties. 
    The impact of physical pooling functions is demonstrated with molecular properties calculated from quantum mechanical computations.
    We also compare our results to the recent set2set pooling approach. 
    We recommend using sum pooling for the prediction of properties that depend on molecular size and compare pooling functions for properties that are molecular size-independent. 
    Overall, we show that the use of physical pooling functions significantly enhances generalization. 
\end{abstract}

\section*{Keywords}
graph convolutional  neural networks, pooling function, physics-informed machine learning, property prediction

\section{Introduction}
Graph neural networks (GNNs) are emerging for end-to-end learning of molecular properties~\cite{Kearnes.2016, Niepert2016, Hamilton2017,Duvenaud.2015} in a broad variety of applications including chemical engineering~\cite{Schweidtmann2020GNNIgnitionQuality,li2021introducing,rittig2022activity,rittig2022octaneFuelDesign}, (quantum) chemistry~\cite{Gilmer.05.04.2017,Yang.2019,Schutt.2017,Wu.2018,back2019convolutional} and the prediction of physical~\cite{Coley.2017} and crystal properties~\cite{Chen.2019,xie2018crystal}. 
Although GNNs are flexible models for end-to-end learning, we show that their pooling function needs to be carefully selected because wrong decisions can lead to unphysical GNNs that are more prone to overfitting. 
GNNs take molecular graphs as inputs and represent atoms by nodes and bonds by edges. 
In addition, atoms and nodes are characterized by corresponding feature vectors. 
Most commonly, GNN architectures are based on message passing neural networks (MPNNs)~\cite{Gilmer.05.04.2017}. 
In MPNNs, the node feature vectors are updated through a series of message passing procedures of neighboring nodes. 
Each of these sequential message passings corresponds to the graph convolutional layers of the GNN. 
Then, the resulting node feature vectors are combined into a molecular fingerprint vector through pooling. 
This molecular fingerprint is finally mapped to molecular properties of interest by feedforward artificial neural networks (ANNs). 
We show that the selection of the pooling function, which combines the feature vectors of all atoms into the molecular fingerprint, is critical for the GNN's performance. \newline \indent 
The vast majority of previous works does not select pooling functions based on physical understanding. 
In the previous literature, common pooling functions are mean, sum, and max pooling~\cite{Wu.2021}. 
Most previous works use sum pooling~\cite{Xu.01.10.2018, Coley.2017, Yang.2019, Lu.2019}, while a few use mean pooling~\cite{morritfey.19, Shindo.8312019}. 
Moreover, typically the same pooling function is used for a range of different properties. 
We argue that this can lead to unphysical GNNs and result in unnecessary errors. 
An illustrative example is the molecular weight that is given by the sum of the atom weights. 
In this case, using mean pooling in a standard GNN would lead to an unphysical architecture that cannot learn the correct underlying physics because the molecular mass cannot be computed as an average of atom weights. 
In contrast, selecting the sum pooling function according to the underlying physics enables the GNN to learn a meaningful model. \newline \indent 
Some researchers circumvent the issue of selecting pooling functions by introducing flexible models for pooling such as set2set~\cite{Vinyals.11192015}, DiffPool~\cite{Ying.2018}, or SortPool~\cite{Zhang.2018}. 
For example, the set2set approach employs a long short-term memory (LSTM) architecture designed for unordered and size-variant input sets~\cite{Vinyals.11192015}.
The authors of DiffPool propose a hierarchical GNN structure that progressively coarsens the input graph in each layer by aggregating clusters of nodes until a single graph representation is obtained~\cite{Ying.2018}. 
SortPool arranges the learned node representation in a consistent ordered tensor which is then truncated or extended to a user-defined fixed size~\cite{Zhang.2018}. 
Similarly, GNNs with a large number of convolutional layers combine node information through convolutions and thus reduce the importance of pooling functions~\cite{morritfey.19}. 
Advanced pooling methods have also been applied in molecular property prediction. 
For instance, Gilmer et al. (2017)~\cite{Gilmer.05.04.2017} applied the set2set method for learning various molecular properties from the QM9 data set~\cite{Ruddigkeit.2012, Ramakrishnan.2014} and achieved state-of-the-art accuracies on all target properties compared to other GNN models at the time of publication. 
However, the additional flexibility typically results in larger data requirements, higher model variance, and the risk of overfitting. 
In other words, the selection of physical pooling functions over flexible model architectures for pooling can be understood as enforcing a hybrid model structure, which is known to reduce the data demand~\cite{Psichogios.1992,fiedler2008local,schweidtmann2021machine}. \newline \indent
A few recent studies emphasize the importance of the choice of pooling functions for property prediction. 
Xu et al. (2018)~\cite{Xu.01.10.2018} examine sum, mean, and max pooling and conclude that sum pooling is more powerful than mean and max pooling since it can better distinguish different graph structures. 
Pronobis et al. (2018)~\cite{Pronobis.2018} state that decomposition of molecules into atom-wise contributions combined with a property-suitable pooling function works better for ``extensive properties''. 
Other works use property-specific pooling functions, where mean/set2set or sum pooling is applied to ``intensive'' or ``extensive'' properties, respectively~\cite{Schutt.12112018, Schutt.2018, Gubaev.2018, Ye.12162019, Liu.2021}. 
Overall, the physical selection of pooling functions is somewhat contradictory in the previous literature and the terms ``intensive'' or ``extensive'' have been used colloquially and not in their thermodynamic sense. 
Also, a comparison of different pooling functions on the prediction and generalization capabilities of GNNs against a physical background has not been conducted yet, hence there is no guide for selecting suitable pooling functions based on physical knowledge in the literature. \newline \indent
We evaluate GNN pooling functions against the underlying physical nature of the learned properties. 
We analyze the impact of physical pooling functions on molecular properties learned from the common QM9 data set~\cite{Ruddigkeit.2012, Ramakrishnan.2014} and demonstrate their superior performance. 

\section{Materials and methods}
We first describe the graph representation of molecules and then briefly introduce the general GNN architecture used for property prediction.
Finally, we provide physical insight into the learned properties and use the insight to design physical GNN architectures.

\subsection{Molecular graph} 
\label{sec:PhysicalGNN_Molecular_Graph}
Molecules can be described as molecular graphs with nodes $v,w\in V$ representing atoms and edges $e_{vw} \in E$ representing bonds.
Each atom is described by a feature vector $\textbf{f}^V(v)$, containing atom information, e.g., atom mass or orbital hybridization. 
Similarly, each bond is described by a bond feature vector $\textbf{f}^{E}(e_{vw})$ that contains information on the bond type, e.g., single or double bond. 
Commonly, for organic molecules the hydrogen (H) atoms are omitted and replaced by the hydrogen count as a node feature~\cite{Todeschini.2000};  this results in reducing the complexity of molecular graphs and therefore reducing data demand.

\subsection{Graph neural network} 
\label{sec:PhysicalGNN_GNNs}
\begin{figure}
	\centering
	\includegraphics[width=\textwidth]{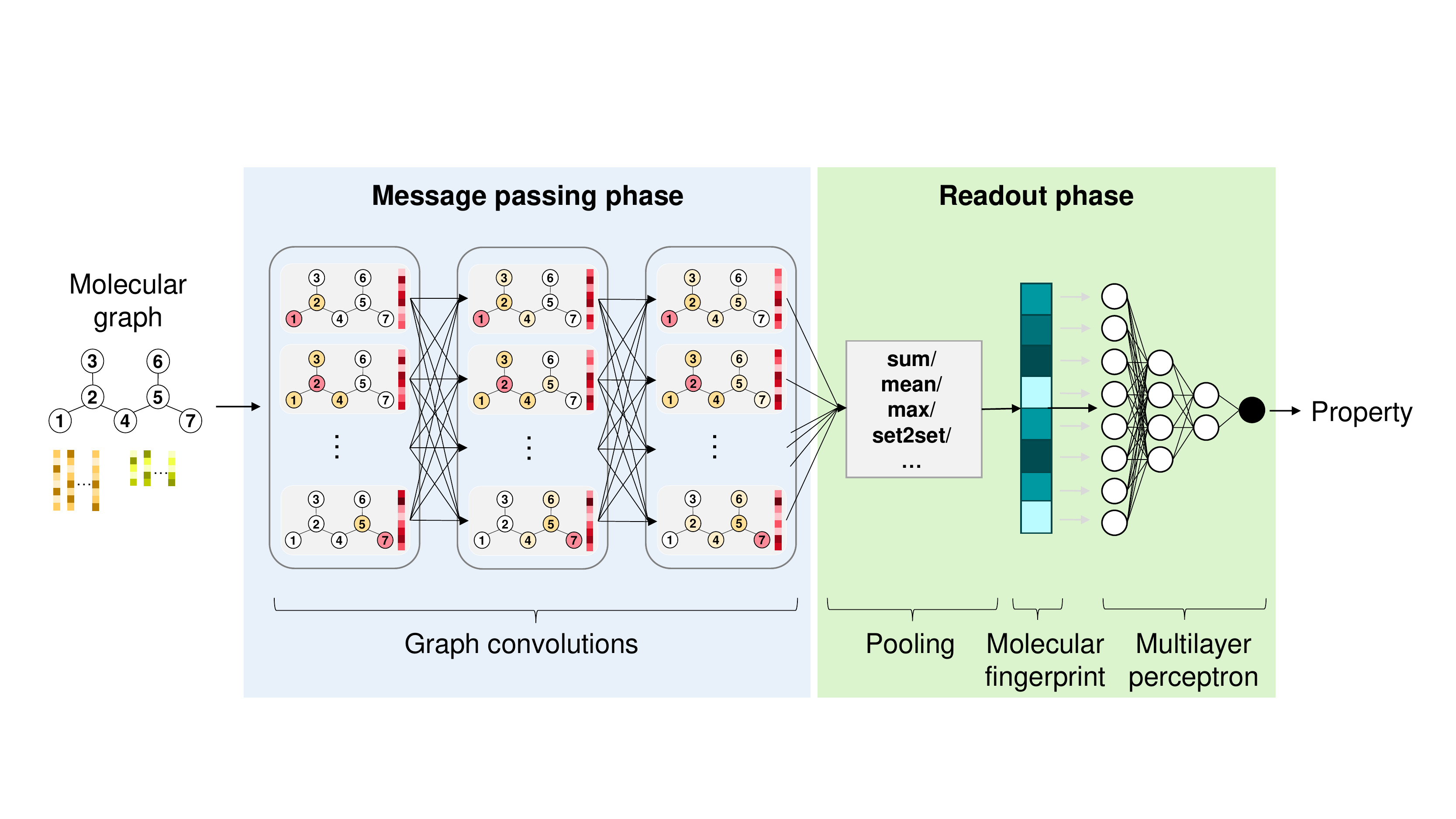}
	\caption{Illustration of the GNN structure highlighting the message passing and readout phases.}
	\label{fig:PhysicalGNN_GNN_structure}
\end{figure}

GNNs exhibit two phases~\cite{Gilmer.05.04.2017} as shown in Figure~\ref{fig:PhysicalGNN_GNN_structure}: 
(i) message passing phase and 
(ii) readout phase. \newline \indent
To initialize the message passing phase, each node $v\in V$ is assigned a state vector $\textbf{h}_{v}^{l=0}$ initialized by the respective node feature vector~\cite{Gilmer.05.04.2017}.
Then, the state vector of the nodes in layer $l$ are updated with information from their neighboring nodes $w \in N(v)$ along edges $e_{vw}$:
\begin{equation*}
	\textbf{h}_v^l~=~U_l \left( \textbf{h}_{v}^{l-1}, \sum_{w \in N(v)} M_{l} \left( \textbf{h}_{v}^{l-1}, \textbf{h}_w^{l-1}, \textbf{f}^{E}(e_{vw}) \right) \right), 
\end{equation*}
where $U_l(\cdot)$ and $M_l(\cdot)$ respectively denote the state update function and the message function in layer $l$. 
This message passing procedure is repeated $L$ times until each node state vector $\textbf{h}_{v}^{L}$ includes information about its local environment. 
This iterative message passing corresponds to the stacking of $L$ graph convolutional layers. \newline \indent
We consider a standard GNN including edge features in the message passing phase, also known as 1-GNN~\cite{morritfey.19, Hamilton.17.09.2017}. 
The 1-GNN uses the following message passing function:
\begin{equation*}
	\textbf{h}_{v}^{l} = \sigma \left(\theta_{v}^l \cdot \textbf{h}_{v}^{l-1} + \sum_{w \in N(v)} \text{ANN}_{\theta_e^l}(\textbf{f}^{E}(e_{vw})) \cdot \textbf{h}_w^{l-1} \right), 
\end{equation*}{}
where $\sigma$ indicates an activation function, $\theta_v^l$ denotes a parameter matrix, and $\text{ANN}_{\theta_e^l}$ denotes a feedforward ANN mapping the respective feature vectors $\textbf{f}^{E}(e_{vw})$ of the edges $e_{vw}$ connecting node $v$ with its neighbors to a parameter matrix $\theta_e^l$, referred to as edge feature network.

In the readout phase, the final state vectors of the nodes $\textbf{h}_{v}^{L}$ are combined into a graph state vector $\textbf{h}_{G}$ by a pooling function. 
The pooling function is necessary for molecular property prediction because the number of atoms usually differs between different molecules. 
This leads to a varying number of atom feature vectors that need to be combined to the molecular fingerprints. 
Thus, the pooling function combines a varying number of final state vectors for the nodes into a single graph state vector. 
In the context of molecular property prediction, the literature commonly refers to $\textbf{h}_{G}$ as the molecular fingerprint.
This molecular fingerprint $\textbf{h}_{G}$ is finally fed into a feed-forward ANN for the prediction of molecular properties, $\hat{\textbf{p}} = \text{MLP}(\textbf{h}_{G})$. \newline \indent
The molecular fingerprint is given by the pooling function $f_p(\cdot)$ that depends on the final state vectors of the nodes $\textbf{h}_{v}^{L}$ with $v \in V$: 
\begin{equation*}
	\textbf{h}_{G} = f_p\left( \{\textbf{h}_{v}^{L} \mid v \in V \} \right)
\end{equation*}
Common choices for $f_p$ are the $\sumWord$, $\mean$, and $\max$ functions.
An alternative pooling function is the set2set method~\cite{Vinyals.11192015} which can capture more complex relationships between different atomic contributions~\cite{Gilmer.05.04.2017, Schutt.2018} by employing a long short-term memory~(LSTM) model~\cite{Vinyals.11192015}. 
After $T$ steps of the following iterative computation, the molecular fingerprint is obtained by $\textbf{h}_{G} = \textbf{q}_{t=T}^{*}$ with
\begin{equation*}
	\begin{gathered}
		\textbf{q}_{t}^{*} = \textbf{q}_{t} \mathbin\Vert \textbf{r}_{t} \label{eqn:PhysicalGNN_set2set} \\
		\textbf{q}_{t} = \text{LSTM}\left(\textbf{q}_{t-1}^{*}\right) \\
		\textbf{r}_{t} = \sum_{v} \textbf{a}_{v,t} \cdot \textbf{h}_{v}^{L} \\
		\textbf{a}_{v,t} = \text{softmax}\left( \textbf{h}_{v}^{L} \cdot \textbf{q}_{t} \right)
	\end{gathered}
\end{equation*}
where $\textbf{q}_{t}$ is a query vector for iteration $t$ providing information about the previous attention readout vector $\textbf{r}_{t}$ from the memories, $\textbf{a}_{v,t}$ is an attention vector resulting from averaging the attention of a node $v$ by applying the softmax function, $\textbf{r}_{t}$ is the attention readout, similar to the simple pooling method with sum, and $\textbf{q}_{t}$ is a concatenation ($\mathbin\Vert$) of the current query vector and the attention readout. 
The vector $\textbf{q}_{t-1}^{*}$ is initialized at $t=0$ by $\textbf{q}_{-1}^{*}=\textbf{0}$. \newline \indent
Recent GNNs incorporate physical knowledge into message passing.
Over the last years, multiple MPNN architectures have been proposed that integrate physical knowledge to the message passing scheme, e.g., SchNet~\cite{Schutt.2018}, PhysNet~\cite{Unke.2019}, DimeNet~\cite{Klicpera.06.03.2020}, MXMNet~\cite{zhang2020molecular}. 
This includes the incorporation of directional information, such as interatomic distances and angles between atom pairs, into the message function $M_l(\cdot)$ modeling the interactions of atoms.
We consider MXMNet that utilizes physical-driven message passing while preserving computational efficiency~\cite{zhang2020molecular}. 
Within MXMNet, two message passing schemes are applied. 
In a global message passing scheme, information between atoms with a global cutoff distance $d_g$ is exchanged. 
Further, a local message passing is applied to exchange information between atoms with a local cutoff distance $d_l$ with $d_g > d_l$. 
This local cutoff distance represents the connectivity of atoms that are connected by chemical bonds. 
The architecture further enables to transfer of information between atom representations in the global and local message passing by including a cross layer mapping. 
In the readout step, the learned atom-wise representations are subsequently pooled by the sum operator for molecular property prediction. 

\subsection{Physical insight}
\label{sec:PhysicalGNN_Physical_Insight}
The prediction of molecular properties by decomposing molecules into atomic contributions has a long history in chemical research. 
According to Bonchev~\cite{Bonchev.1991}, the first investigations into properties of molecules with additive characteristics in terms of atomic contributions were carried out in the 1850s. 
Later, quantitative structure-property relationship~(QSPR) and group additivity methods were developed based on the additive character of atoms or functional groups within a molecule~\cite{Katritzky.1995, Benson.1969, Gani.1991, JOBACK.1987}.
Yet, not every molecular property exhibits purely additive effects. \newline \indent
In thermodynamics, macroscopic properties are categorized as intensive or extensive~\cite{QuantitiesUnitsSymbols}. 
A system property is extensive if it scales linearly with the mass (and as such ``extent'') of the system; examples are the mass or volume. 
In contrast, intensive properties do not change with the system mass. 
Note that sometimes thermodynamicists also distinguish between intensive (e.g., temperature and pressure) and specific (extensive quantity divided by volume, e.g., density) properties~\cite{stephan2013thermodynamik}, but we will not. 
We transfer these concepts to molecules and distinguish between \textit{molecular size-independent} and \textit{molecular size-dependent} properties. 
Molecular size-dependent properties scale with the number of atoms in a molecule. 
For example, the molecular weight is determined by how many atoms of which type are present in a molecule. 
In contrast, there exist molecular size-independent properties that do not scale with the number of atoms in a molecule, e.g., the highest occupied molecular energy level (HOMO)~\cite{Schutt.2018}. 
Moreover, some properties of a substance, e.g., activity or toxicity, are mostly influenced by certain functional groups or structural fragments. 
Note that this dependency on molecular size does not necessarily correspond to the formal definition of intensive or extensive properties in a thermodynamic sense because the former is considered at a microscopic atom-based molecular level, not at a macroscopic mass-based system level. 
For example, the molar enthalpy with the unit J/mol is an intensive property. 
On a molecular level, the enthalpy of atomization $H_{298,atom}$ with the unit J/mol, is a molecular size-dependent property describing the amount of energy needed to break up a molecule into all of its single atoms at room temperature and fixed pressure~\cite{Gilmer.05.04.2017}. \newline \indent
Schütt et al.~\cite{Schutt.2018} consider the QM9 properties dipole moment ($\mu$), isotropic polarizability ($\alpha$), electronic spatial extent ($\text{R}^2$), zero point vibrational energy (ZPVE), heat capacity at 298.15K ($\text{C}_{\text{v},298}$), atomization energy at 0K ($\text{U}_{\text{0,atom}}$), atomization energy at 298.15K ($\text{U}_{\text{298,atom}}$), enthalpy of atomization at 298.15K ($\text{H}_{\text{298,atom}}$), free energy of atomization at 298.15K ($\text{G}_{\text{298,atom}}$) as ``extensive''. Other properties are the highest occupied molecular orbital ($\epsilon_{\text{HOMO}}$), lowest unoccupied molecular orbital energy level ($\epsilon_{\text{LUMO}}$), and HOMO-LUMO gap ($\Delta\epsilon$).
For some of these properties, physical dependencies on molecular size are known. 
For example, Miller and Savchik~\cite{Miller79} developed a semi-empirical approach for the prediction of isotropic polarizabilities as a sum of atomistic contributions that depend on their hybridization states based on theoretical calculations already in 1979. 
Each nonlinear molecule has $3N-6$ vibrational degrees of freedom ($3N-5$ for linear ones), $N$ being its number of atoms. 
Each degree of freedom has a ZPVE proportional to its frequency $\nu \propto \sqrt{f/m}$. 
Here, $m$ is the reduced mass of the parts of a molecule that vibrate with respect to each other and $f$ is the force constant of this vibration. 
Hence, ZPVE is a molecular size-dependent property to first order but the frequencies of vibrations that include large fractions of a molecule decrease with increasing molecular size. 
This effect should be learned by the GNN. 
Similar relations apply to the heat capacity, enthalpy, and entropy contributions with the minor complication that terms dependent on the molecular mass (translation) and the moment of inertia (rotation) arise for some of the contributions~\cite{Atkins2011QM}. 
Atomization energies and enthalpies include essentially sums of contributions of all bonds, which may be non-local in the case of conjugated bonds, and are hence molecular size-dependent properties as well. 
The electronic spatial extent is determined mainly by the shapes of the orbitals in very small molecules and is closely related to the radius of gyration for large molecules, which may even depend on the solvent for large molecules such as polymers and may scale with a fractal exponent in this case. \newline \indent
The dipole moment ($\mu$) is a particularly interesting property. 
Even though formally the molecule size enters the dipole moment equation, in most molecules local functional groups determine $\mu$, and depending on orientation they can even weaken each other. 
In particular, one or a few strongly polar groups dominate the dipole moment that can then be written as the sum of the individual dipole moments vectors. 
Thus, we classify the dipole moment as molecular size independent. 
The prediction of dipole moment is expected to be challenging for conventional GNNs because long-range orientational relations between the polar groups may need to be learned by a model, e.\,g.,~for describing the difference between the polar ortho- and the unpolar para-benzoquinone. \newline \indent
Similarly, the relations are complex for energies of the HOMO, the LUMO, and their difference (i.e., the HOMO-LUMO-gap). 
Depending on the type of molecule, these orbitals may be quite localized to a certain functional group and thus independent of molecular size in some molecules. 
However, they may also be delocalized in other molecules and thus dependent on molecular size for small and medium-sized molecules but converge to a limit for large molecules as can be seen, e.g., from the H\"uckel model for conjugated double bonds \cite{Atkins2011QM}. 
Hence this property may be particularly challenging for a GNN model. 

\section{Results and discussion}
In order to demonstrate the relevance of the pooling function in the readout phase, two case studies are conducted and discussed below.
First, we consider the illustrative prediction of the molecular weight. 
Then, we consider the prediction of twelve quantum mechanical properties collected in the QM9 data set. 

\subsection{Hyperparameters and implementation}
Our implementations are based on the models in PyTorch Geometric developed by Fey \& Lenssen~\cite{Fey.362019}. 
For our case study, we combine each mean, sum, and max as well as set2set pooling with the 1-GNN. 
The hyperparameters of the models are selected based on our experience from our previous work on predicting fuel properties~\cite{Schweidtmann2020GNNIgnitionQuality}. 
The molecular graphs have the following features encoded as a one-hot vector: (node) atom type, is aromatic, is in ring, hybridization (e.g., $sp, sp^2,sp^3$) hydrogen count, (edge) bond type, conjugated, and stereo.
The 1-GNN comprises three graph convolutional layers with hidden dimension size of 64. 
To map the molecular fingerprint ($\textbf{h}_{G}$) to the property ($\hat{p}$) of interest, we use multilayer perceptrons (MLPs), $\hat{p} = \text{MLP}(\textbf{h}_{G})$. 
The MLPs constitute four layers with \#1: 64, \#2: 32 \#3: 16, \#4: 1 neurons when mean, sum, or max pooling is used. 
When the set2set pooling is used, the MLP layers have \#1: 128, \#2: 64, \#3: 32, \#4: 1 neurons, because the output vector of the set2set method is twice its input size. 
For the set2set method, we set the number of processing steps $T$ to 3. \newline \indent
We additionally test mean, sum, max pooling with MXMNet. 
We use the implementation and default hyperparameters with batch size 128 and global cutoff distance 5 as it was provided by the authors of MXMNet~\cite{zhang2020molecular}, cf.~\cite{zhang2020github}.

\subsection{Illustrative case study: Molecular weight}
\label{sec:PhysicalGNN_molecular_weight_case_study}
To illustrate the importance of physical pooling functions on a simple example, we learn the molecular weight of alkanes.
To compose the data set, we obtain about 2,300 alkanes from $\text{C}_1\text{H}_4$ to $\text{C}_{60}\text{H}_{122}$ from the PubChem database~\cite{Kim.2016} and compute their molecular weight using RDKit~\cite{rdkit}. \newline \indent
For illustration, we split this case study into two steps. 
Firstly, 1-GNNs with sum, mean, and max pooling functions are trained, validated, and tested on corresponding data sets with alkanes with up to 30 C-atoms. 
Secondly, the trained GNNs are tested against an external data set containing alkanes with more than 30 C-atoms. 
Thus, the generalization capabilities of the 1-GNNs are tested against extrapolated data. \newline \indent 
The training of the 1-GNNs is repeated ten times for each pooling function with a maximum number of 500 epochs.
The initial data set with up to 30 C-atoms is randomly split into 80\% training, 10\% validation, and 10\% test sets for each run. 
Since we consider alkanes, we choose the attributes of the nodes in the molecular graph to include the hydrogen count only and do not use edge attributes, hence we replace the edge feature network in the message passing of the 1-GNN by a (learned) parameter matrix that is the same for all edges. \newline \indent 
\begin{figure}
	\begin{subfigure}[c]{0.49\textwidth}
		\centering
		\includegraphics[width=\textwidth]{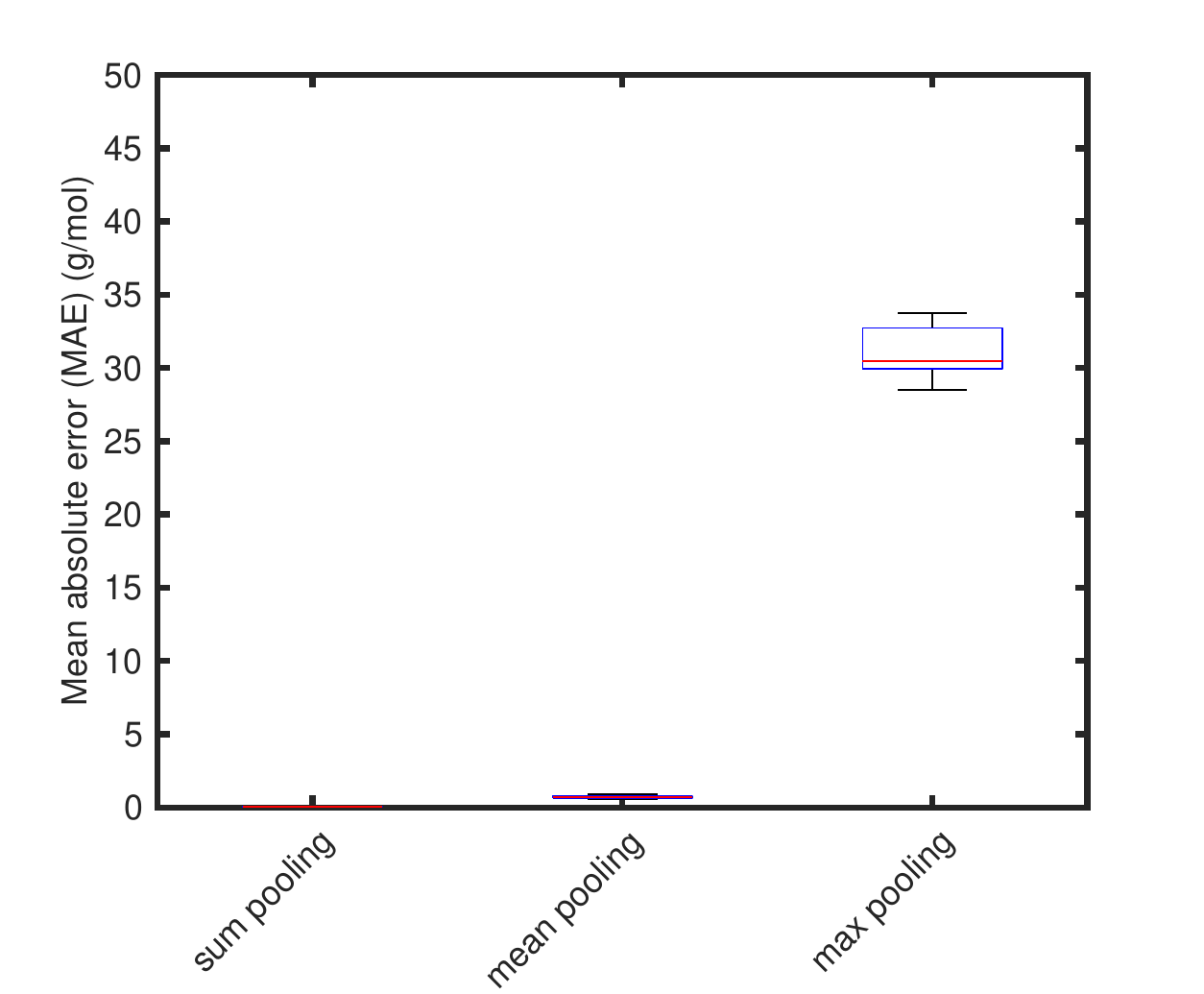}
		\subcaption{}
	\end{subfigure}
	\begin{subfigure}[c]{0.49\textwidth}
		\centering
		\includegraphics[width=\textwidth]{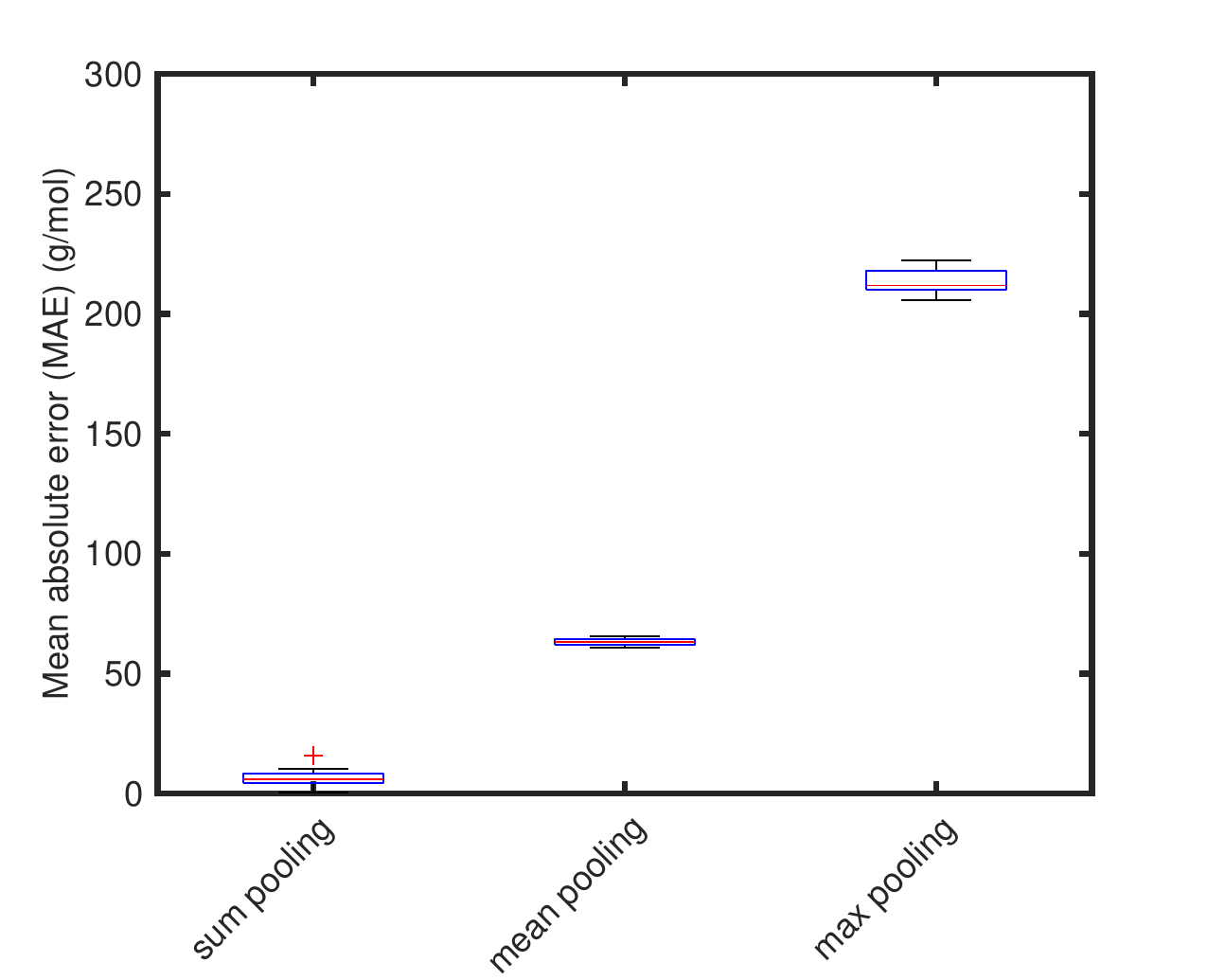}
		\subcaption{}
	\end{subfigure}
	\caption{Mean absolute error in g/mol for test of the 1-GNN with different pooling functions, namely: sum, mean, max. (a) test data set of alkanes with up to 30 C-atoms, (b) external data set of alkanes with 35 up to 60 C-atoms. Results are for ten independent training runs, each with 500 periods.}
	\label{fig:PhysicalGNN_Alkane}
\end{figure} 
Figure~\ref{fig:PhysicalGNN_Alkane} shows the test set performance of the 1-GNN with different pooling functions. 
Figure~\ref{fig:PhysicalGNN_Alkane} (a) illustrates the test results for the data set of alkanes with up to 30 C-atoms. 
As expected, the sum pooling leads to the best performance on the test data set with an average mean absolute error of $0.06~\text{g/mol}$ as it captures the molecular size-dependent character of the molecular weights. 
In contrast, mean pooling leads to an average mean absolute error $0.7~\text{g/mol}$. 
Max pooling even leads to an average absolute error in the order of $31~\text{g/mol}$. \newline \indent 
Figure~\ref{fig:PhysicalGNN_Alkane} (b) shows the mean absolute error on the test data sets of alkanes with more than 30 C-atoms. 
The 1-GNN with sum pooling leads to an average absolute error of $6.5~\text{g/mol}$. 
In contrast, the 1-GNN with mean pooling leads to an average error of $63~\text{g/mol}$ and the 1-GNN with max pooling leads to an average error of $213~\text{g/mol}$. \newline \indent 
The results clearly show that the sum pooling function, which was selected based on our physical insight, performs better than the unphysical mean and max pooling functions for the prediction of the molecular weight. 
In particular, the sum pooling outperforms the unphysical pooling functions significantly when extrapolating the model. 
These results support our theoretical expectations that GNNs with unphysical pooling functions are more prone to overfitting. 
Notably, the mean absolute error of the GNN with mean pooling is much smaller on the test set with alkanes with less than 30 C-atoms compared to the test set with alkanes with more than 30 C-atoms. 
This result indicates that the selection of pooling functions based on the performance on a standard validation or test set could also be error-prone. 

\subsection{Physicochemical properties}
\label{sec:PhysicalGNN_QM9_case_study}
We analyze the importance of physical pooling functions for a variety of relevant properties. 
In addition, we compare our results to a more complex set2set readout function and explore the MXMNet~\cite{zhang2020molecular} architecture. \newline \indent
We use the QM9 data set to train our models~\cite{Ruddigkeit.2012, Ramakrishnan.2014}. 
The experimental setup is twofold. 
First, the 1-GNNs with sum, mean, max, and set2set pooling functions are trained, validated, and tested on randomly selected subsets of the whole QM9 data set. 
This approach assesses the interpolation capabilities of the respective pooling functions. 
Second, we train, validate, and test the 1-GNN and MXMNet on the QM9 data excluding molecules with exactly 9 heavy atoms. 
These models are then tested against molecules with 9 heavy atoms from the QM9 data set. 
This approach is chosen to assess the generalization capabilities of the models in terms of extrapolation ability. 
For each training run, the data set is randomly split into 80\% training, 10\% validation, and 10\% test sets. 
The training is stopped after 300 and 900 periods for the 1-GNN and MXMNet, respectively. 
The mean absolute error for the test set is reported based on the period with the lowest validation error. 

\subsubsection{Interpolation}
\label{sec:PhysicalGNN_Physical_GNN_Interpolation}
The results for testing sum, mean, max, and set2set pooling function on the whole QM9 data set are summarized in Table~\ref{tab:PhysicalGNN_standard_QM9}. 
Overall, the results indicate that physically meaningful pooling functions lead to favorable performances on the QM9 data set for interpolation. 
It can be observed that for all molecular size-dependent properties, the 1-GNN with sum pooling significantly outperforms the 1-GNNs with mean and max pooling. 

For the molecular size-independent properties, we do not observe a superior performance of one pooling function; all pooling functions result in similar accuracies.

\begin{table}
	\centering
	\caption{Mean absolute errors averaged over three training runs for testing 1-GNN with different pooling functions, sum, mean, max, set2set against QM9 target properties. Properties are categorized into molecular size-independent (``m.~size-ind.'') and molecular size-dependent (``m.~size-dep.'') according to~\cite{Schutt.2018}. Errors of the best pooling function are bold type.}
	\begin{tabular}{crrrrrr}
		\toprule
		\multicolumn{3}{c}{\multirow{2}[4]{*}{\textbf{Target}}} & \multicolumn{4}{c}{\textbf{Pooling}} \\
		\cmidrule{4-7}   \multicolumn{3}{c}{}                                                              & sum               & mean              & max               & set2set   \\
		\midrule
		\multirow{9}{*}{m. size-dep.}   & $\alpha$                          & {$\text{a}_0^3$}              & \textbf{0.301}    & 0.469             & 0.482             & 0.583     \\
		                                & $\text{R}^2$                      & {$\text{a}_0^2$}              & \textbf{21.5}     & 25.0              & 24.1              & 24.1      \\
	                                	& $\text{ZPVE}$                     & {meV}                          & \textbf{9.39}  & 24.16           & 26.99           & 21.96   \\
		                                & $\text{C}_{\text{v},298}$         & {${\text{cal}}/\text{mol K}$} & \textbf{0.149}    & 0.203             & 0.197             & 0.197     \\
		                                & $\text{U}_{0,\text{atom}}$        & {eV}                          & \textbf{0.117}    & 0.345             & 0.357             & 0.437     \\
	                                    & $\text{U}_{298,\text{atom}}$      & {eV}                          & \textbf{0.117}    & 0.363             & 0.366             & 0.386     \\
		                                & $\text{H}_{298,\text{atom}}$      & {eV}                          & \textbf{0.123}    & 0.329             & 0.383             & 0.390     \\
		                                & $\text{G}_{298,\text{atom}}$      & {eV}                          & \textbf{0.112}    & 0.293             & 0.328             & 0.313     \\
		\hline
		\multirow{3}{*}{m. size-ind.}   & $\mu$                             & {Debye}                       & 0.452             & 0.456             & \textbf{0.449}    & 0.472     \\
		& $\epsilon_{\text{HOMO}}$          & {meV}                          & 92.8            & 93.1            & \textbf{92.6}   & 133.7    \\
		                                & $\epsilon_{\text{LUMO}}$          & {meV}                          & \textbf{93.1}   & 93.7            & 93.8            & 93.7    \\
		                                & $\Delta\epsilon$                  & {eV}                          & 0.1322             & 0.1336           & \textbf{0.1283}   & 0.1323    \\
		\bottomrule
	\end{tabular}
	\label{tab:PhysicalGNN_standard_QM9}
\end{table}

\subsubsection{Generalization with 1-GNN architecture}
\label{sec:PhysicalGNN_Physical_GNN_Extrapolation}
In order to analyze the generalization capability of the pooling functions on the QM9 data set, we train the 1-GNNs only on molecules with up to 8 heavy atoms, i.e., a maximum number of 8 C, N, O, F atoms. 
Then, we test the prediction accuracy on an internal test set, i.e., containing molecules with up to 8 heavy atoms, and also on a data set with the remaining QM9 molecules that have exactly 9 heavy atoms. The latter test set therefore tests the extrapolation capability of the derived GNNs. 
The results of the extrapolation are summarized in Table~\ref{tab:PhysicalGNN_standard_QM9_1_GNN_extrapolation}. \newline \indent 
The interpolation performance of the models (indicated in black in Table~\ref{tab:PhysicalGNN_standard_QM9_1_GNN_extrapolation}) is similar to that of the previous models trained on the whole QM9 data set (cf. Table~\ref{tab:PhysicalGNN_standard_QM9}). 
Notably, the absolute errors increase for some properties, e.g., atomization energy, as the training set is much smaller. 
The QM9 data set contains about 108,000 molecules with 9 atoms and about 22,000 molecules with 1 to 8 heavy atoms. \newline \indent
For extrapolation (indicated in blue in Table~\ref{tab:PhysicalGNN_standard_QM9_1_GNN_extrapolation}), we find that sum pooling performs much better compared to mean, max, and set2set pooling on all tested molecular size-dependent properties. 
For the molecular size-independent properties we observe that the sum pooling does not outperform the other pooling functions anymore. 
Rather, mean and max pooling perform slightly better compared to sum pooling. 
Notably, the set2set method does not improve the accuracy compared to sum, mean, and max pooling for any property.
\begin{table}
	\centering
	\caption{Mean absolute errors averaged over three independent training runs for testing 1-GNN with different pooling functions, sum, mean, max, and set2set, against: (black) QM9 data set excluding molecules with 9 heavy atoms, (blue) only molecules with 9 heavy atoms of the QM9 data set. The errors of the best pooling function are bold type. Units are equivalent to those in Table~\ref{tab:PhysicalGNN_standard_QM9}.}
	\begin{tabular}{rrrrr}
		\toprule
		\multirow{2}[4]{*}{\textbf{Target}} & \multicolumn{4}{c}{\textbf{1-GNN}} \\
		\cmidrule{2-5}                                  & sum                               & mean                              & max                                   & set2set                   \\
		\midrule
		\multirow{2}[0]{*}{$\alpha$}                    & \textbf{0.302}                    & 0.816                             & 0.727                                 & 1.164                     \\
		                                                & \textcolor{rwth}{\textbf{1.385}}  & \textcolor{rwth}{8.445}           & \textcolor{rwth}{8.654}               & \textcolor{rwth}{7.835} \\
		\multirow{2}[0]{*}{$\text{R}^2$}                & \textbf{17.9}                     & 24.9                              & 23.3                                  & 29.4 \\
		                                                & \textcolor{rwth}{\textbf{83.6}}   & \textcolor{rwth}{236.0}           & \textcolor{rwth}{229.5}               & \textcolor{rwth}{227.0} \\
		\multirow{2}[0]{*}{$\text{ZPVE}$}               & \textbf{0.0123}                   & 0.0561                            & 0.0544                                & 0.0547 \\
		                                                & \textcolor{rwth}{\textbf{0.0288}} & \textcolor{rwth}{0.3661}          & \textcolor{rwth}{0.3127}              & \textcolor{rwth}{0.4508} \\
		\multirow{2}[0]{*}{$\text{C}_{\text{v},298}$}   & \textbf{0.158}                    & 0.312                             & 0.291                                 & 0.309 \\
		                                                & \textcolor{rwth}{\textbf{0.650}}  & \textcolor{rwth}{3.509}           & \textcolor{rwth}{3.360}               & \textcolor{rwth}{3.129} \\
		\multirow{2}[0]{*}{$\text{U}_{0,atom}$}         & \textbf{0.180}                    & 0.816                             & 0.766                                 & 0.773 \\
		                                                & \textcolor{rwth}{\textbf{1.365}}  & \textcolor{rwth}{8.082}           & \textcolor{rwth}{8.100}               & \textcolor{rwth}{9.613} \\
		\multirow{2}[0]{*}{$\text{U}_{298,atom}$}       & \textbf{0.171}                    & 0.759                             & 0.757                                 & 0.631 \\
	                                                	& \textcolor{rwth}{\textbf{1.239}}  & \textcolor{rwth}{8.368}           & \textcolor{rwth}{8.117}               & \textcolor{rwth}{8.220} \\
		\multirow{2}[0]{*}{$\text{H}_{298,atom}$}       & \textbf{0.181}                    & 0.724                             & 0.802                                 & 0.706 \\
		                                                & \textcolor{rwth}{\textbf{1.197}}  & \textcolor{rwth}{8.275}           & \textcolor{rwth}{8.125}               & \textcolor{rwth}{10.424} \\
		\multirow{2}[1]{*}{$\text{G}_{298,atom}$}       & \textbf{0.167}                    & 0.690                             & 0.705                                 & 0.757 \\
		                                                & \textcolor{rwth}{\textbf{1.142}}  & \textcolor{rwth}{7.671}           & \textcolor{rwth}{7.363}               & \textcolor{rwth}{8.751} \\
		\hline
		\multirow{2}[1]{*}{$\mu$}                       & 0.469                             & 0.465                             & \textbf{0.454}                        & 0.501                     \\
		& \textcolor{rwth}{0.588}           & \textcolor{rwth}{0.572}           & \textcolor{rwth}{\textbf{0.569}}      & \textcolor{rwth}{0.611}   \\
		\multirow{2}[0]{*}{$\epsilon_{HOMO}$}           & 0.110                             & \textbf{0.109}                    & 0.112                                 & 0.127 \\
		                                                & \textcolor{rwth}{0.142}           & \textcolor{rwth}{\textbf{0.140}}  & \textcolor{rwth}{0.142}               & \textcolor{rwth}{0.158} \\
		\multirow{2}[0]{*}{$\epsilon_{LUMO}$}           & 0.116                             & 0.114                             & \textbf{0.110}                        & 0.118 \\
		                                                & \textcolor{rwth}{0.179}           & \textcolor{rwth}{0.171}           & \textcolor{rwth}{\textbf{0.167}}      & \textcolor{rwth}{0.173} \\
		\multirow{2}[0]{*}{$\Delta\epsilon$}            & 0.161                             & 0.156                             & \textbf{0.150}                        & 0.156 \\
		                                                & \textcolor{rwth}{0.226}           & \textcolor{rwth}{0.219}           & \textcolor{rwth}{\textbf{0.208}}      & \textcolor{rwth}{0.219} \\
		\bottomrule
	\end{tabular}
	\label{tab:PhysicalGNN_standard_QM9_1_GNN_extrapolation}%
\end{table}%

\subsubsection{Generalization with MXMNet architecture}
We also analyze the influence of the pooling function on the MXMNet GNN model~\cite{zhang2020molecular}. 
The MXMNet model reached state-of-the-art performance on several prediction tasks in QM9~\cite{zhang2020molecular}.
MXMNet includes directional information in its message passing process. 
For the readout step, sum pooling is applied in the original MXMNet model.
We compare MXMNet performance with three different pooling functions: sum, mean, and max.
Similar to the 1-GNN, we test MXMNet trained on molecules of QM9 with up to 8 heavy atoms against an internal test set and an external test set, i.e., extrapolating to molecules with 9 heavy atoms.
The results are summarized in Table~\ref{tab:PhysicalGNN_standard_QM9_MXMNet_extrapolation}. \newline \indent
The interpolation performance of the MXMNet (indicated in black in Table~\ref{tab:PhysicalGNN_standard_QM9_MXMNet_extrapolation}) is highly favorably compared to the 1-GNN architecture, as expected. 
The MXMNet with sum pooling outperforms the other pooling approaches for all molecular size-dependent properties. 
This is in agreement with our expectations and previous observations on the 1-GNN architecture. 
For the molecular size-independent properties, we observe that sum, mean, and max pooling perform very similarly. 
Notably, the extrapolation performance of the MXMNet architecture (indicated in blue in Table~\ref{tab:PhysicalGNN_standard_QM9_MXMNet_extrapolation}) also significantly outperforms the 1-GNN architecture on all properties but $R^2$. 
\newline \indent
The generalization results follow the same pattern as our previous observations. 
Again, we observe a significant advantage of sum pooling for all molecular size-dependent properties.
For the molecular size-independent properties, we obtain similar performance of the pooling functions for LUMO.
For $\mu$, sum pooling performs only slightly better than mean and max pooling. 
For the HOMO and the HOMO-LUMO gap, however, mean and max pooling outperform sum pooling by a factor of more than 3.
This demonstrates that also sum pooling can promote overfitting and thus prevent generalization in case of size-independent properties. 
Notably, the extrapolation error of the MXMNet is much larger for the unphysical pooling function compared to the extrapolation error of the simpler 1-GNN with the same pooling function. 
This indicates that the selection of physical pooling functions could be more important for more complex models. 

\begin{table}
	\centering
	\caption{Mean absolute errors averaged over three independent training runs for testing MXMNet with different pooling functions, sum, mean, max, against: (black) QM9 data set excluding molecules with 9 heavy atoms, (blue) only molecules with 9 heavy atoms of the QM9 data set. Error of best pooling function are bold type. Units are equivalent to those in Table~\ref{tab:PhysicalGNN_standard_QM9}.}
	\begin{tabular}{rrrrr}
		\toprule
		\multirow{2}[4]{*}{\textbf{Target}} & \multicolumn{3}{c}{\textbf{MXMNet}} \\
		\cmidrule{2-4}                                  & sum                                   & mean                              & max                                   \\
		\midrule
		\multirow{2}[0]{*}{$\alpha$}                    & \textbf{0.0781}                       & 0.1331                            & 0.1347                                 \\
		                                                & \textcolor{rwth}{\textbf{0.1887}}     & \textcolor{rwth}{1.2853}          & \textcolor{rwth}{0.4569}               \\
		\multirow{2}[0]{*}{$\text{R}^2$}                & \textbf{1.78}                         & 2.55                              & 2.77                                  \\
		                                                & \textcolor{rwth}{\textbf{104.84}}     & \textcolor{rwth}{199.26}          & \textcolor{rwth}{182.34}               \\
		\multirow{2}[0]{*}{$\text{ZPVE}$}               & \textbf{0.00144}                      & 0.00370                           & 0.00278                                  \\
		                                                & \textcolor{rwth}{\textbf{0.00226}}    & \textcolor{rwth}{0.04492}         & \textcolor{rwth}{0.00807}               \\
		\multirow{2}[0]{*}{$\text{C}_{\text{v},298}$}   & \textbf{0.0325}                       & 0.0503                            & 0.0515                                 \\
		                                                & \textcolor{rwth}{\textbf{0.0891}}     & \textcolor{rwth}{2.2484}          & \textcolor{rwth}{0.6094}               \\
		\multirow{2}[0]{*}{$\text{U}_{0,atom}$}         & \textbf{0.0111}                       & 0.0728                            & 0.0710                                 \\
		                                                & \textcolor{rwth}{\textbf{0.0265}}     & \textcolor{rwth}{1.4095}          & \textcolor{rwth}{0.5047}              \\
		\multirow{2}[0]{*}{$\text{U}_{298,atom}$}       & \textbf{0.0114}                       & 0.0729                            & 0.0720                                 \\
	                                                	& \textcolor{rwth}{\textbf{0.0265}}     & \textcolor{rwth}{1.2072}          & \textcolor{rwth}{0.5025}              \\
		\multirow{2}[0]{*}{$\text{H}_{298,atom}$}       & \textbf{0.0115}                       & 0.0723                            & 0.0724                                 \\
		                                                & \textcolor{rwth}{\textbf{0.0265}}     & \textcolor{rwth}{1.5855}          & \textcolor{rwth}{0.5039}              \\
		\multirow{2}[1]{*}{$\text{G}_{298,atom}$}       & \textbf{0.0122}                       & 0.0662                            & 0.0677                                 \\
		                                                & \textcolor{rwth}{\textbf{0.0271}}     & \textcolor{rwth}{2.1216}          & \textcolor{rwth}{0.5009}              \\
		\hline
		\multirow{2}[1]{*}{$\mu$}                       & \textbf{0.0892}                       & 0.0981                            & 0.1203                                \\
		& \textcolor{rwth}{\textbf{0.1551}}              & \textcolor{rwth}{0.1708}          & \textcolor{rwth}{0.1997}      \\
		\multirow{2}[0]{*}{$\epsilon_{HOMO}$}           & 0.0516                                & \textbf{0.0460}                   & 0.0550                                  \\
		                                                & \textcolor{rwth}{0.2273}              & \textcolor{rwth}{\textbf{0.0687}} & \textcolor{rwth}{0.0743}                \\
		\multirow{2}[0]{*}{$\epsilon_{LUMO}$}           & \textbf{0.0366}                       & 0.0384                            & 0.0449                                  \\
		                                                & \textcolor{rwth}{\textbf{0.0700}}     & \textcolor{rwth}{0.0707}          & \textcolor{rwth}{0.0816}                \\
		\multirow{2}[0]{*}{$\Delta\epsilon$}            & 0.0766                                & \textbf{0.0729}                   & 0.0783                                 \\
		                                                & \textcolor{rwth}{0.3896}              & \textcolor{rwth}{\textbf{0.1192}} & \textcolor{rwth}{0.1263}       \\
		\bottomrule
	\end{tabular}
	\label{tab:PhysicalGNN_standard_QM9_MXMNet_extrapolation}%
\end{table}%

\section{Conclusion}
GNNs have emerged as a promising deep learning technique for end-to-end molecular property prediction in chemical engineering. 
The selection of pooling functions in GNNs for property prediction should be based on physical knowledge because incorrect pooling functions can promote overfitting and weaken generalization. 
We identify the dependency of the learned property on the molecular size as key property for the selection of the pooling function: When a property is molecular size-dependent, the sum pooling function should be used. When a property is molecular size-independent, sum pooling can lead to poor generalization. We recommend to compare sum, mean, and max pooling functions for size-independent properties. 

Our computational results support this hypothesis showing that physical GNN architectures generalize better than unphysical architectures. 
In future research, the physical selection of pooling functions should always be considered when predicting molecular properties with GNNs.

\section*{Acknowledgements}
Supported by the German Research Foundation (DFG) within the framework of the Excellence Strategy of the Federal Government and the Länder - Cluster of Excellence 2186 ``The Fuel Science Center'' (ID390919832). 
Simulations were performed with computing resources granted by RWTH Aachen University under projects thes0682 and rwth0731. 
MD received funding from the Helmholtz Association of German Research Centers.

\section*{Data and software availability}

Our implementations are based on the models in PyTorch Geometric developed by Fey \& Lenssen~\cite{Fey.362019}. 
The model implementation is available at \url{https://git.rwth-aachen.de/avt-svt/public/graph_neural_network_for_fuel_ignition_quality} under Eclipse Public License 2.0 (cf.~\cite{Schweidtmann2020GNNIgnitionQuality}). 
The MXMNet implementation is provided by the authors of MXMNet~\cite{zhang2020molecular}, cf.~\cite{zhang2020github}.
We use the QM9 data set to train our models~\cite{Ruddigkeit.2012, Ramakrishnan.2014}. 

\section*{Authors contributions}
\textbf{Artur M. Schweidtmann}: Conceptualization, Methodology, Validation, Formal analysis, Investigation, Writing - Original Draft, Writing - Review \& Editing, Visualization. 
\textbf{Jan G. Rittig}: Methodology, Software, Validation, Formal analysis,  Investigation, Data Curation, Writing - Original Draft, Writing - Review \& Editing, Visualization.	
\textbf{Jana M. Weber}: Methodology, Validation, Formal analysis, Writing - Review \& Editing.		
\textbf{Martin Grohe}: Methodology, Validation, Formal analysis, Writing - Review \& Editing.			
\textbf{Manuel Dahmen}: Methodology, Validation, Formal analysis, Writing - Review \& Editing, Supervision.
\textbf{Kai Leonhard}: Methodology, Validation, Formal analysis, Writing - Review \& Editing.			
\textbf{Alexander Mitsos}: Methodology, Validation, Formal analysis, Resources, Writing - Review \& Editing, Supervision, Project administration, Funding acquisition.

\bibliographystyle{abbrv}       
\bibliography{Literatur}   

\end{document}